\documentclass[10pt,twocolumn,letterpaper]{article}

\usepackage[pagenumbers]{cvpr} 
\usepackage[accsupp]{axessibility}  

\usepackage[accsupp]{axessibility} 
\usepackage{graphicx}
\usepackage{amsmath}
\usepackage{amssymb}
\usepackage{booktabs}

\usepackage{pifont} 
\newcommand{\cmark}{\ding{51}}%
\newcommand{\xmark}{\ding{55}}%

\usepackage{graphicx}
\usepackage{booktabs}
\usepackage{multirow}
\usepackage{makecell}
\usepackage{caption}
\usepackage{threeparttable}
\usepackage{algorithm} 
\usepackage{algorithmic} 
\usepackage{listings}
\usepackage{bm}
\usepackage[dvipsnames]{colortbl}
\usepackage[dvipsnames]{xcolor} 
\usepackage{hhline}
\usepackage{arydshln}
\usepackage{array,etoolbox}
\usepackage{tcolorbox}

\definecolor{KleinBlue}{rgb}{0.0, 0.129, 0.7}

\newcommand{\methodname}{BOLT}

\definecolor{cvprblue}{rgb}{0.21,0.49,0.74}
\usepackage[pagebackref,breaklinks,colorlinks,allcolors=cvprblue]{hyperref}

\def\paperID{3937} 
\def\confName{CVPR}
\def\confYear{2025}

\title{BOLT: \underline{Bo}ost \underline{L}arge Vision-Language Model Without \underline{T}raining \\  for {L}ong-form Video Understanding}

\author{
Shuming Liu \qquad
Chen Zhao\thanks{Corresponding author.} \qquad
Tianqi Xu \qquad
Bernard Ghanem
\and
King Abdullah University of Science and Technology (KAUST)
}

\begin{document}
\maketitle
\begin{abstract}
Large video-language models (VLMs) have demonstrated promising progress in various video understanding tasks. However, their effectiveness in long-form video analysis is constrained by limited context windows. Traditional approaches, such as uniform frame sampling, often inevitably allocate resources to irrelevant content, diminishing their effectiveness in real-world scenarios. 
In this paper, we introduce \textbf{BOLT}, a method to \textbf{BO}ost \textbf{L}arge VLMs without additional \textbf{T}raining through a comprehensive study of frame selection strategies. First, to enable a more realistic evaluation of VLMs in long-form video understanding, we propose a multi-source retrieval evaluation setting. Our findings reveal that uniform sampling performs poorly in noisy contexts, underscoring the importance of selecting the right frames. 
Second, we explore several frame selection strategies based on query-frame similarity and analyze their effectiveness at inference time. Our results show that \textbf{inverse transform sampling} yields the most significant performance improvement, increasing accuracy on the Video-MME benchmark from 53.8\% to 56.1\% and MLVU benchmark from 58.9\% to 63.4\%. Our code is available at \url{https://github.com/sming256/BOLT}.
\end{abstract}
    
\section{Introduction}
\label{sec:intro}

Recently, large Video-Language Models (VLMs) have emerged as powerful tools for video understanding~\cite{maaz2023video,lin2023mm,zhang2023simple,lin2023video,jin2024chat,ren2024timechat,Liu_2024_CVPR}. These models leverage the synergy between visual and linguistic modalities through large language models, enabling them to capture complex relationships and contextual cues over time. They have demonstrated remarkable advancements in tasks such as video captioning~\cite{yangVid2SeqLargeScalePretraining2023, chenShareGPT4VideoImprovingVideo2024, wuDIBSEnhancingDense2024} and video question answering~\cite{kimImageGridCan2024, minMoReVQAExploringModular2024, maaz2023video}.
However, large VLMs face significant challenges in processing long videos containing thousands of frames due to the sheer volume of video tokens. For instance, LLaVA-OneVision~\cite{liLLaVAOneVisionEasyVisual2024a} utilizes 196 tokens per frame, meaning that without reduction, an hour-long video at 30 FPS would require 20 million tokens. Given the limited context length of the underlying language models and computational constraints~\cite{yuFrameVoyagerLearningQuery2024,shenLongVUSpatiotemporalAdaptive2024}, \textbf{video token reduction} has become essential for enabling large VLMs to efficiently handle video understanding tasks.

\begin{figure}[t]
\centering
\includegraphics[width=1.0\linewidth]{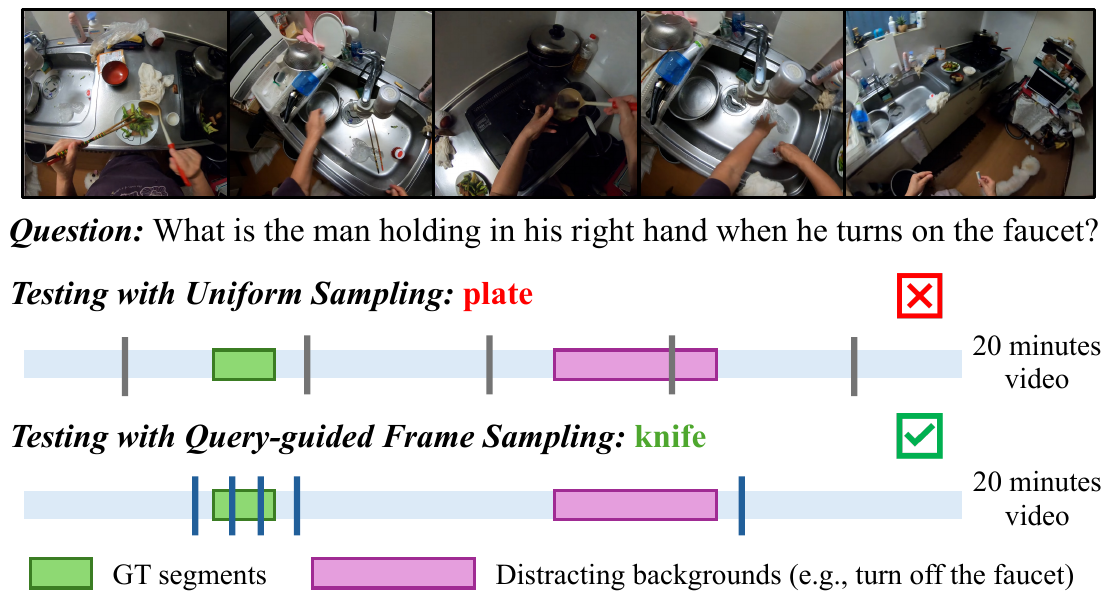}
\vspace{-18pt}
\caption{Given a long video, conventional VLMs adopt uniform sampling during inference, which may divert VLM’s focus due to the distracting backgrounds. In contrast, our query-guided frame sampling can identify and prioritize question-relevant frames, boosting the VLM's performance without requiring fine-tuning.}
\label{fig:intro}
\vspace{-12pt}
\end{figure}

To reduce video tokens, previous works have primarily explored strategies from two perspectives. The first is \textit{spatial compression}, which reduces the number of tokens per frame, as employed by methods such as MovieChat~\cite{songMovieChatDenseToken2024} and SlowFast-LLaVA~\cite{xuSlowFastLLaVAStrongTrainingFree2024}. This paper focuses on the second perspective, \textit{temporal selection}, which involves selecting a subset of frames from raw videos. Most prior works have relied on uniform frame sampling, in which a fixed number of frames are evenly selected from the video \cite{ataallah2024minigpt4,li2023videochat,ye2024mplug}. Although simple and efficient, this approach often fails in complex real-world scenarios because it assumes that all frames contribute equally to content understanding, which may hold true for short, query-trimmed videos but not for long, unedited ones. For example, instructional or surveillance videos usually contain lengthy segments irrelevant to the given query. Sampling frames from these segments not only consumes the token budget but may also distract the VLM from pertinent information, thereby impairing its performance~\cite{li2023videochat,shuVideoXLExtraLongVision2024}.

Some recent methods, such as Goldfish \cite{ataallahGoldfishVisionLanguageUnderstanding2024a}, LongVU \cite{shenLongVUSpatiotemporalAdaptive2024}, and Frame-Voyager~\cite{yuFrameVoyagerLearningQuery2024}, have started to explore adaptive temporal selection strategies by using visual-text similarity to identify relevant frames. However, these methods require additional training on existing VLMs, demanding substantial computational resources and extensive fine-tuning data. For instance, Frame-Voyager~\cite{yuFrameVoyagerLearningQuery2024} requires 32 H100 GPUs for training. This raises a crucial question: \textit{Can we boost the performance of existing large VLMs without additional training by designing better temporal selection methods?} This paper provides an affirmative answer to this question through two key explorations: how \textit{selecting the right frames} matters and how to \textit{select the frames right}.

\textit{First}, we investigate the importance of \textit{selecting the right frames} for long-form video understanding using two newly designed evaluation setups. (1) We utilize the MultiHop-EgoQA dataset~\cite{chen2024grounded}, which provides question-related segments, to examine the performance of large VLMs with and without access to ground-truth video segments. (2) To further study the impact of frame selection in real-world scenarios containing diverse background clips in longer videos, we propose a multi-source retrieval evaluation setting by extending the videos via internal retrieval in the existing dataset. The results show that selecting irrelevant frames, \textit{i.e.}, query-unrelated ones, significantly degrades the performance on VQA tasks — a common issue encountered with uniform sampling.

\textit{Then}, we explore how to \textit{select the frames right} by introducing several query-aware frame selection approaches, including top-\textit{K} selection, watershed grouping, and inverse transform sampling. All these approaches are applied only at test time, without any retraining of the VLMs. We find that inverse transform sampling consistently improves VLM performance, particularly in the multi-source retrieval evaluation scenario. Using this frame selection strategy, we can \textbf{BO}ost the performance of existing \textbf{L}arge VLMs without additional \textbf{T}raining. We thus name our method \textbf{BOLT}.

Our main contributions can be summarized as:

\begin{itemize}
    \item We demonstrate that selecting the right frames significantly impacts large VLMs' test-time performance for long-form video understanding. We validate this through experiments on the MultiHop-EgoQA dataset~\cite{chen2024grounded}, and by introducing multi-source retrieval evaluation setup, which better reflects real-world video understanding challenges.
    \item We explore various training-free frame selection strategies and propose leveraging inverse transform sampling to select query-relevant frames to improve VQA performance without requiring model retraining.
    \item Our method achieves consistent improvements on five benchmarks, \textit{e.g.}, from 53.8\% to 56.1\% on the Video-MME~\cite{fu2024video}, and from 58.9\% to 63.4\% on MLVU~\cite{zhou2024mlvu}. 
\end{itemize}

\section{Related Works}
\label{sec:related_works}

\subsection{Large Vision-Language Models for Video}

Transformer-based large language models (LLMs) have revolutionized natural language processing~\cite{dubeyLlama3Herd2024, openaiGPT4TechnicalReport2024a}. Researchers have extended LLMs to handle multiple modalities by incorporating multimodal inputs, typically visual content such as images and videos~\cite{zhuMiniGPT4EnhancingVisionLanguage2023, liLLaVAOneVisionEasyVisual2024a} to make large vision-language models (VLMs). These models leverage large-scale pre-training on extensive data to learn rich representations that capture intricate relationships between visual and textual modalities.
Building upon these, large VLMs have significantly advanced various video understanding tasks such as video captioning~\cite{yangVid2SeqLargeScalePretraining2023, chenShareGPT4VideoImprovingVideo2024, wuDIBSEnhancingDense2024}, video question answering~\cite{kimImageGridCan2024, minMoReVQAExploringModular2024, maaz2023video}, and temporal reasoning~\cite{wuZeroShotLongFormVideo2024, qianMomentorAdvancingVideo2024}. Recent surveys~\cite{tangVideoUnderstandingLarge2024,liang2024survey} comprehensively review the architectures, training strategies, and applications of VLMs in video understanding, highlighting both their capabilities and the challenges they address.

Despite these successes, large VLMs encounter significant challenges when processing videos due to the substantial amount of video frames, the limited context length of language models~\cite{wu2024longvideobench}, and the ``lost in the middle'' problem~\cite{liuLostMiddleHow2023a}. These constraints necessitate efficient frame selection strategies to identify and retain essential information from videos without overwhelming the model. Thus, developing effective frame selection methods is critical to further enhance VLM performance in video understanding tasks.

\subsection{Video Token Reduction in Large VLMs}

Reducing video tokens is essential for making large VLMs computationally feasible. The most common and straightforward approach is uniform frame sampling, which selects frames without considering their individual importance to the task. Unfortunately, this method may omit crucial frames needed for understanding complex or long videos, thereby limiting reasoning performance. Several strategies have been proposed in the literature to make token reduction more adaptive and task-specific.

\vspace{4pt}
\noindent\textbf{Token Merging.} One approach is to merge information within or across frames to form a compact yet comprehensive representation. MovieChat~\cite{songMovieChatDenseToken2024} implements short-term and long-term memory mechanisms to integrate and compress frames from a long video, ensuring minimal information loss. LLaMA-VID~\cite{liLLaMAVIDImageWorth2023} encodes each frame into separate context and content tokens, whereas LongVU~\cite{shenLongVUSpatiotemporalAdaptive2024} leverages cross-modal queries and inter-frame dependencies to reduce redundancy while preserving visual details. However, token merging approaches risk oversimplifying fine-grained information, potentially preventing the model from capturing subtle differences between frames.

\vspace{4pt}
\noindent\textbf{Query-Based Frame Sampling.} To overcome limitations associated with uniform sampling, recent similarity-based methods select frames based on their relevance to specific queries or tasks. Goldfish~\cite{ataallahGoldfishVisionLanguageUnderstanding2024a} addresses video QA by segmenting long videos into shorter clips and retrieving relevant clips using cosine similarity between query embeddings and captions. Other approaches learn optimal frame selection strategies through additional training. For instance, Frame-Voyager~\cite{yuFrameVoyagerLearningQuery2024} learns informative frame combinations by training on a dataset that ranks frames based on prediction losses from a VLM~\cite{chen2023videollm}. Although potentially more effective, such learnable sampling methods require additional training and substantial computational resources, limiting their practicality.

Unlike the above-mentioned methods, in this work, we propose a training-free frame selection strategy to enhance the performance of existing large VLMs at inference time.

\section{Revisiting Large VLMs}

Consider a video $V$ with $T$ frames $\{f_1, f_2, \ldots, f_T\}$ and an associated query $Q$ (system prompts omitted for brevity). The VQA task aims to generate an answer $A$ that accurately addresses the query based on the content of $V$. The response format depends on the query type, encompassing both open-ended and multiple-choice answers.

State-of-the-art VLMs typically address this task using the following pipeline: First, due to computational constraints and the video's high frame rate, a subset of $N$ frames $\{f'_1, f'_2, \ldots, f'_N\}$ is selected, where $N \ll T$. In common implementations, $N$ is often set to 16 or 32 frames. Next, each selected frame $f'_i$ is processed through a vision encoder $E_v$ to obtain visual embeddings: $v_i = E_v(f'_i)$, where $v_i \in \mathbb{R}^{M \times C}$. Here, $C$ denotes the embedding dimension, and $M$ represents the number of tokens per frame.

Following visual encoding, a projection module maps visual embeddings $v_i$ into the language embedding space, ensuring dimensional compatibility with textual tokens. For simplicity, we omit the projector from subsequent notation. The visual embeddings are concatenated with the question’s textual embeddings $E_t(Q)$, forming an input sequence processed by an LLM to generate the answer auto-regressively:
\begin{equation}
A = \text{LLM}([v_1; v_2; \ldots; v_N; E_t(Q)]).
\end{equation}

Note that the context-length limitation $L$ of the above language model, \textit{eg.,}~16K, constrains the total number of tokens, including visual and textual embeddings:
\begin{equation}
\sum_{i=1}^N \text{len}(v_i) + \text{len}(E_t(Q)) \leq L.
\end{equation}

Unlike image understanding tasks that process a single frame, this constraint makes frame selection crucial for effective video understanding, as the model must select frames judiciously within the limited context window.

\section{\methodname}

In this section, we first perform an empirical analysis to validate the importance of frame selection. This analysis incorporates annotated ground-truth segments from untrimmed videos and a challenging multi-source retrieval condition. Subsequently, we compare four different frame selection strategies, finding that inverse transform sampling can effectively prioritize question-relevant frames.

\subsection{Impact of Ground Truth Segments}
In VQA tasks, particularly those involving long-form video understanding, many queries pertain to specific temporal segments within hour-long untrimmed videos. We hypothesize that effectively retrieving query-relevant segments significantly reduces the complexity of reasoning for VLMs. To validate this hypothesis, we leverage existing VQA benchmarks that provide temporal interval annotations.

\begin{table}[t]
\small
\setlength\tabcolsep{4pt}
\centering
\caption{\textbf{Demonstration of the importance of selecting the right frames} (\textit{i.e.}, GT segments) on MultiHop-EgoQA dataset. All models use 8 frames as input. We report the average sentence similarity as the evaluation metric.}
\vspace{-4pt}
\begin{tabular}{lrr}
\toprule
\textbf{Frame Selection} & \textbf{LLaVA-NeXT-Video-7B} & \textbf{InternVL2-8B}  \\
\midrule
Uniform          & 61.7   & 71.8 \\
GT segments      & \textbf{62.8} (+1.1)   & \textbf{72.5} (+0.7) \\
w.o. GT segments & 59.2 (-2.5) & 69.1 (-2.7) \\
\bottomrule
\end{tabular}
\label{table:anaylysis_multihop}
\vspace{-8pt}
\end{table}

Specifically, we utilize MultiHop-EgoQA~\cite{chen2024grounded}, a recently introduced long-form egocentric benchmark. This dataset provides temporal annotations of relevant intervals for each video-question pair, such as $\{(t_{s1},t_{e1}), (t_{s2},t_{e2}), \ldots\}$. To examine the significance of frame selection, we conduct experiments under three conditions: (1) \textit{Uniform sampling (baseline)}: Frames are uniformly sampled from the entire video. (2) \textit{Ground-truth segment-only sampling}: Frames are uniformly sampled only from question-relevant ground-truth segments. (3) \textit{Complement-set sampling}: Frames are uniformly sampled only from segments outside the ground-truth intervals. The results are summarized in Table~\ref{table:anaylysis_multihop}.

We evaluate pretrained models InternVL2-8B~\cite{chen2024internvl} and LLaVA-NeXT-Video-7B~\cite{zhang2024llavanextvideo} using 8 sampled frames, reporting average sentence similarity as the evaluation metric (higher is better). Experimental results show that, compared with uniform sampling, excluding irrelevant context enables VLMs to better focus on informative visual features for question answering. Conversely, when only irrelevant frames are provided, the generated predictions exhibit lower similarity to ground-truth answers.

\begin{figure*}[t]
\centering
\includegraphics[width=0.97\linewidth]{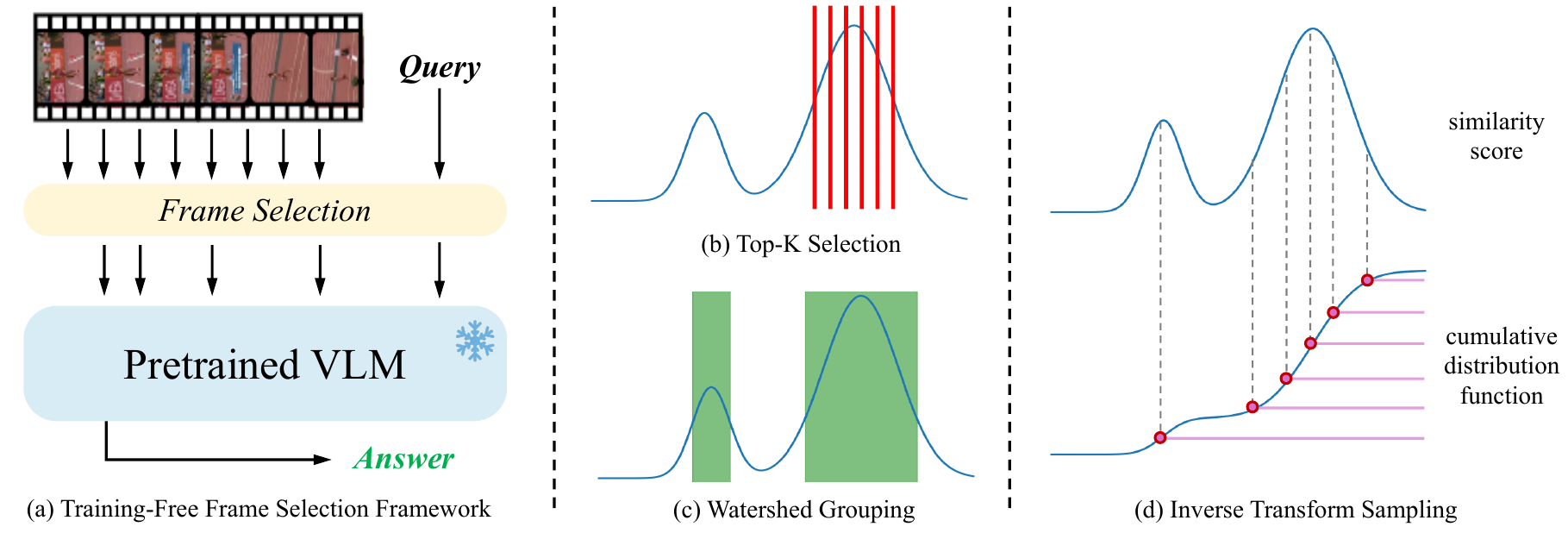}
\vspace{-10pt}
\caption{\textbf{(a) Proposed training-free frame selection framework.} Without exhaustive fine-tuning of VLMs, their performance can be improved by selecting query-related frames. \textbf{(b) Top-K selection} often yields redundant frames with limited diversity. \textbf{(c) Watershed grouping} may overlook broader temporal context. \textbf{(d) Inverse transform sampling} can prioritize relevant frames while preserving selection diversity, achieving stronger performance on benchmarks, particularly for videos with noisy backgrounds. }
\label{fig:method}
\vspace{-10pt}
\end{figure*}

\subsection{Multi-Source Retrieval Evaluation}
\label{sec:multi_source}

\begin{table}[t]
\small
\centering
\caption{\textbf{Multi-source retrieval evaluation on Video-MME.} $K$ is set to 4 in this experiment. Using the `right' frames (GT segments) significantly outperforms uniformly sampling frames.}
\vspace{-4pt}
\begin{tabular}{lr}
\toprule
\textbf{Frame Selection Strategy} & \textbf{LLaVA-OneVision-7B} \\
\midrule
GT segments (16 frames) & 56.85 \\
\midrule
Uniform (64 frames) & 52.11 (-4.74) \\
Uniform (16 frames) & 47.78 (-9.07) \\
\midrule
Blind test & 40.85 \\
\bottomrule
\end{tabular}
\label{table:analysis_videomme}
\vspace{-12pt}
\end{table}

The above preliminary analysis demonstrates the significance of frame selection in VQA tasks by directly evaluating the models' capability on ground-truth segments. However, extending this analysis to general VQA benchmarks, such as Video-MME~\cite{fu2024video}, is constrained by the absence of temporal segment annotations. To overcome this limitation and assess frame selection efficacy under more challenging conditions, we introduce a novel multi-source retrieval evaluation setting.

The proposed multi-source retrieval evaluation employs a systematic construction procedure. First, $K$ distinct videos are sampled from the benchmark and concatenated along their temporal dimension, with brief black-frame transitions inserted between adjacent videos. During video selection, we ensure that the sampled videos have similar durations yet differ significantly in their visual content, such as different categories or domains. This concatenation generates a pseudo-long-form video that effectively simulates real-world continuous video content. Next, we randomly select a question $Q$ from the associated question sets of these videos and evaluate the model's response accuracy using the concatenated video, deviating from the conventional trimmed-video evaluation paradigm.

Our proposed evaluation framework offers distinct advantages over standard evaluation protocols. Primarily, it better approximates real-world scenarios in which models must identify relevant content within extended video sequences, such as lengthy instructional or surveillance footage. Furthermore, it imposes more rigorous reasoning demands on VLMs, providing a more comprehensive assessment of their capabilities. We implement this evaluation framework on the Video-MME dataset, and the preliminary results are presented in Table~\ref{table:analysis_videomme}.

The experimental results reveal several critical insights. Most notably, uniform sampling exhibits substantial performance degradation under proposed evaluation framework, indicating its fundamental inadequacy in handling long video content containing substantial query-irrelevant segments. Despite increasing the input frame count to 64 frames, accuracy still exhibits a significant decline from 56.85\% to 52.11\%. With a fixed frame count of 16 frames, performance further deteriorates to 47.78\%, representing a notable 9.07\% reduction compared to the baseline. This result is even close to the blind test result, in which no visual information is provided. These findings underscore the importance of intelligent frame selection strategies for VQA tasks, particularly when processing long video sequences.

\subsection{Selecting the Frames Right} 

Based on the above analysis, we emphasize that it is essential for VLMs to select frames relevant to the query. Next, we discuss how to select those frames effectively, and introduce several frame selection strategies designed to improve the identification and prioritization of question-relevant frames without requiring model fine-tuning.

\vspace{4pt}
\noindent\textbf{Uniform Sampling.} We first revisit conventional uniform sampling, which selects frames at fixed intervals: $f'_i = f_{\lfloor i \times T/N \rfloor}$, where $i \!\in\! \{1, \ldots, N\}$. Although uniform sampling is computationally efficient, it neglects the query context by implicitly assuming a uniform distribution of information throughout the video. Since uniform sampling remains widely used during both training and testing in current state-of-the-art VLMs, we adopt this as our baseline.

\vspace{4pt}
\noindent\textbf{Top-\textit{K} Selection.}
To identify query-relevant frames, an intuitive approach is to leverage text-video similarity and select frames with high similarity scores. A commonly used method is top-\textit{K} sampling, which operates as follows:

First, the question \( Q \) and each frame \( f_i \) are encoded into embeddings using a pretrained contrastive lightweight multimodal network, such as CLIP~\cite{radford2021learning} or SigLIP~\cite{zhai2023sigmoid}. The similarity between the visual embedding of a frame \( E_v(f_i) \) and the textual embedding of the question \( E_t(Q) \) can then be computed, typically using cosine similarity:
\begin{equation}
s(f_i) = \cos(E_t(Q), E_v(f_i)).
\end{equation}

Next, we select the top \( K \) frames with the highest similarity scores. These query-relevant frames are retained and subsequently processed by the VLM. Formally, the selection process is defined as:
\begin{equation}
\mathcal{F}_{\text{top-}K} = \{ f_i \mid s(f_i) \in \text{Top-}K(\{ s(f_j) \}_{j=1}^N) \}.
\end{equation}

Although effective for direct visual matching, top-\textit{K} sampling exhibits notable limitations. As shown in Figure~\ref{fig:method}(b), when applied to long video sequences, top-\textit{K} sampling often yields redundant frames with high visual similarity, potentially missing diverse contextual information crucial for complex queries. Moreover, this strategy may inadequately address temporal dependencies and causal relationships, which require analyzing scene transitions and broader contextual continuity.

\vspace{4pt}
\noindent\textbf{Watershed Grouping.}
The watershed algorithm, originally developed for image segmentation~\cite{haralick1987image}, can be effectively adapted for temporal clustering in video understanding tasks. This approach conceptualizes similarity scores as a topographical landscape, where local extrema define distinct temporal segments.

Formally, given the frame-query similarity scores $s(f_i)$, we first apply a smoothing operation, such as sliding-window average pooling, to mitigate outliers and obtain refined scores:
$s^r_i\!=\!\text{smooth}(s(f_i))$. Subsequently, the watershed algorithm computes the mean similarity score across the temporal sequence: $\overline{s}\!=\!\text{mean}\{s^r_1, s^r_2, \ldots, s^r_T\}$. This mean score $\overline{s}$ serves as a threshold for segmenting frames into distinct peak and valley regions, based on whether $s^r_i > \overline{s}$. Finally, we select representative frames exclusively from the top $K$ highest-scoring peak regions.

The watershed-based approach is particularly effective for long videos, where traditional top-\textit{K} sampling tends to yield redundant frames. By clustering temporally adjacent frames, this method facilitates the extraction of a more diverse subset of frames. Nevertheless, due to its grouping mechanism, this strategy can overlook important temporal context and still retain temporal redundancy, as illustrated in Figure~\ref{fig:method}(c). This limitation poses challenges, especially when addressing queries involving causal or explanatory reasoning (\textit{e.g.}, \textit{why} or \textit{how}).

\begin{table*}[t!]
\centering
\caption{\textbf{Comparison of our proposed \methodname\  with state-of-the-art models on standard benchmarks.} Inverse transform sampling is used. Our training-free approach \methodname\  enables off-the-shelf VLM LLaVA-OneVision to focus on query-related frames, thereby boosting the VQA performance in various settings. It also outperforms other large VLMs in the literature with comparable model scales.}
\vspace{-5pt}
\small
\setlength\tabcolsep{3pt}
\begin{tabular}{lcccccccccccc}
\toprule
\multirow{2}{*}{\textbf{Model}}    & \multirow{2}{*}{\makecell{\textbf{LLM} \\ \textbf{Size}}} & \multirow{2}{*}{\makecell{\textbf{Training} \\ \textbf{Free}}} & \multicolumn{4}{c}{\textbf{Video-MME} \small{(w.o. sub.)}} &  & \multicolumn{2}{c}{\textbf{EgoSchema}} & \multirow{2}{*}{\makecell{\textbf{LongVideo} \\ \textbf{Bench}}} & \multirow{2}{*}{\textbf{MLVU}}  & \multirow{2}{*}{\textbf{NextQA}} \\
\cline{4-7}
\cline{9-10}
                          &               &        & Overall     & Short        & Medium      & Long   &     &     Full & Subset    &                                 \\ 
\midrule
Video-LLaVA~\cite{lin2023video}               & 7B         & \xmark     & 39.9 & 45.3  & 38.0 & 36.2 && -    & -    & 39.1 & 47.3 & -                    \\
VideoChat2~\cite{li2023videochat}               & 7B         & \xmark     & 39.5 & 48.3  & 37.0 & 33.2 && 54.4 & 63.6 & 39.3 & 44.5 & -                       \\
ShareGPT4Video~\cite{chenShareGPT4VideoImprovingVideo2024}       & 8B         & \xmark     & 39.9 & 48.3 & 36.3 & 35.0 && - & -  & 41.8 & 46.4 & -       \\
Chat-UniVi-V1.5~\cite{jin2024chat}             & 7B         & \xmark     & 40.6 & 45.7  & 40.3 & 35.8 && -    & -    & -    & -  & -                 \\
VideoLLaMA2~\cite{chengVideoLLaMA2Advancing2024}                 & 7B         & \xmark     & 47.9 & 56.0  & 45.4 & 42.1 && -    & 53.1 & -        & - & -          \\
Frame-Voyager~\cite{yuFrameVoyagerLearningQuery2024}             & 8B         & \xmark     & 57.5 & 67.3 & 56.3 & 48.9 && - & -  & - & 65.6 & 73.9   \\
LongVU~\cite{shenLongVUSpatiotemporalAdaptive2024}               & 7B         & \xmark     & 60.9 & 64.7 & 58.2 & 59.5 && - & -  & - & 65.4 & - \\
\midrule
LLaVA-OneVision$^{\text{8 frame}}$~\cite{liLLaVAOneVisionEasyVisual2024a}   & 7B   &    & 53.8 & 63.6 & 52.0 & 45.7 && 59.17 & 62.0 & 54.2 & 58.9 & 77.4                 \\
\textbf{LLaVA-OneVision$^{\text{8 frame}}$+\methodname}                     & 7B   & \cmark  & \textbf{56.1} & \textbf{66.8} & \textbf{54.2} & \textbf{47.3} && \textbf{59.23} & \textbf{62.2} & \textbf{55.6} &\textbf{63.4} & \textbf{77.4}    \\
\hdashline
LLaVA-OneVision$^{\text{16 frame}}$~\cite{liLLaVAOneVisionEasyVisual2024a}  & 7B   &   &  56.9  & 68.3 & 54.0 & \textbf{48.2} && 59.49 & 61.4 & 55.7 & 61.2 & 78.1                \\
\textbf{LLaVA-OneVision$^{\text{16 frame}}$+\methodname}                    & 7B   & \cmark  & \textbf{57.8} & \textbf{69.2} & \textbf{56.8} & 47.3  && \textbf{59.86} & \textbf{61.8} & \textbf{57.0} & \textbf{65.8} & \textbf{78.3}      \\
\hdashline
LLaVA-OneVision$^{\text{32 frame}}$~\cite{liLLaVAOneVisionEasyVisual2024a}  & 7B   & & 58.5 & \textbf{70.3}    & 56.6  & 48.8 && 60.36 & 62.2 & 56.4 & 63.2 & 79.4              \\
\textbf{LLaVA-OneVision$^{\text{32 frame}}$+\methodname}                    & 7B   & \cmark  & \textbf{59.9}  & 70.1 & \textbf{60.0} & \textbf{49.6}  && \textbf{60.66} & \textbf{64.0}  & \textbf{59.6} & \textbf{66.8} & \textbf{79.5}     \\
\bottomrule                       
\end{tabular}
\label{table:benchmark_result}
\end{table*}

\vspace{4pt}
\noindent\textbf{Inverse Transform Sampling.} 
Analysis of top-\textit{K} sampling and watershed grouping methods reveals two critical requirements: the selection of query-relevant frames and the preservation of frame diversity. Additionally, contextual information surrounding query-relevant segments should be retained. To meet these requirements without additional model fine-tuning, we propose an adaptive frame selection method based on \textit{inverse transform sampling} (ITS).

ITS is a probabilistic sampling technique that generates samples from a specified probability distribution. When applied to video understanding, ITS leverages frame-query similarity scores to construct a query-guided probability distribution over frames, where the probability of selecting a frame \( f_i \) is proportional to its similarity score \( s_i \). Specifically, we first normalize the similarity scores \( s_i \) across the entire video sequence to the range [0, 1] and refine the normalized score as follows:
\begin{equation}
    s^r_i = \left(\frac{s_i - \min(s)}{\max(s) - \min(s)}\right)^\alpha,
\label{eg:alpha}
\end{equation}
where \(\alpha\) controls the sharpness of the distribution. Smaller values of \(\alpha\) produce a more uniform distribution, increasing frame diversity, while larger values yield a sharper distribution, emphasizing query-relevant frames. 

Next, we construct the cumulative distribution function, \( F(i) \), based on the refined scores:
\begin{equation}
    F(i) = \frac{\sum_{j=1}^{i}s^r_j}{\sum_{j=1}^{N}s^r_j}.
\end{equation}

Finally, we sample \( N \) frames using uniform sampling over the inverse transform of this cumulative distribution function:
\begin{equation}
    f'_i = f_{\arg\min_k\{F(k)\geq i/N\}}, \quad i \in \{1, \ldots, N\},
\end{equation}
where \( f'_i \) represents the \( i \)-th sampled frame.

By employing the above inverse transform sampling, frames with higher relevance to the query have an increased probability of being sampled, while frames with lower relevance retain a non-zero probability. This ensures an effective balance between capturing essential query-related visual content and preserving sufficient contextual and temporal diversity. Additionally, by controlling the distribution sharpness through the parameter \(\alpha\), ITS provides flexible adjustment of the relevance-diversity trade-off. Empirically, we find that setting \(\alpha\) between 2.5 and 3.5 yields satisfactory improvements.

Importantly, ITS requires no additional training and can be seamlessly integrated into existing VLMs during inference. Our experimental results demonstrate that ITS consistently outperforms alternative approaches, achieving a 2.3\% improvement over uniform sampling on the Video-MME benchmark and a 4.5\% improvement on the MLVU benchmark.

\section{Experiments}

\subsection{Experimental Settings}

\noindent \textbf{Datasets.} We evaluate our approach on five widely used VQA benchmarks: Video-MME~\cite{fu2024video}, EgoSchema~\cite{mangalam2023egoschema}, LongVideoBench~\cite{wu2024longvideobench}, MLVU~\cite{zhou2024mlvu}, and NextQA~\cite{xiao2021next}. Among them, the NextQA and EgoSchema datasets contain short videos averaging less than 2 minutes in length, focusing primarily on short-form question answering. In contrast, Video-MME, MLVU, and LongVideoBench include longer videos, ranging from a few minutes to around two hours, thus evaluating models' capabilities in long-form video understanding. Notably, Video-MME consists of 900 videos with 2,700 questions and categorizes videos into short, medium, and long subsets, which we specifically leverage in our ablation studies. Importantly, we evaluate Video-MME without subtitles.

\vspace{4pt} 
\noindent \textbf{Implementation Details.} For computational efficiency, we sample video frames at 1 FPS. To compute the similarity between video frames and textual queries, we use the pretrained CLIP-L/14 model~\cite{radford2021learning} to extract visual and textual embeddings. The hyperparameter \(\alpha\) is set to 2.5 for EgoSchema, LongVideoBench, NextQA, and to 3 for Video-MME and MLVU. We utilize the LMMs-Eval library~\cite{zhang2024lmmsevalrealitycheckevaluation} to measure accuracy for multiple-choice tasks. All experiments are conducted on two A100 GPUs. Additional details can be found in the supplementary material.

\subsection{Benchmark Results}

\begin{table}[t!]
\centering
\caption{\textbf{Our training-free approach \methodname\  boosts various VLMs.} Inverse transform sampling is used on the Video-MME benchmark. All models use 8 frames.}
\vspace{-5pt}
\small
\setlength\tabcolsep{3pt}
\begin{tabular}{lccccccccc}
\toprule
\multirow{2}{*}{\textbf{Model}}   & \multirow{2}{*}{\makecell{\textbf{LLM} \\ \textbf{Size}}} & \multirow{2}{*}{\makecell{\textbf{With} \\ \textbf{\methodname}}}  & \multicolumn{4}{c}{\textbf{Video-MME} \small{(w.o. sub.)}}  \\
\cline{4-7}
                          &             &        & Overall     & Short        & Medium      & Long                            \\ 
\midrule
\multirow{2}{*}{Video-LLaVA}  & \multirow{2}{*}{7B} &  \xmark & 37.6 & 42.7 & 37.1 & 33.0  \\
& & \cmark & \textbf{39.3} & \textbf{45.7} & \textbf{38.0} & \textbf{34.1}  \\
\midrule
\multirow{2}{*}{LongVA}  & \multirow{2}{*}{7B} &  \xmark & 48.5 & 56.7 & 46.6 & 42.3  \\
& & \cmark & \textbf{51.6} & \textbf{59.8} & \textbf{51.0} & \textbf{43.9} \\
\midrule
\multirow{2}{*}{InternVL2}  & \multirow{2}{*}{8B} &  \xmark & 52.6 & 62.4 & 51.1 & 44.2 \\
& & \cmark & \textbf{53.3} & \textbf{65.6} & \textbf{50.4} & \textbf{43.9}  \\
\midrule
\multirow{2}{*}{LLaVA-Video}  & \multirow{2}{*}{7B} &  \xmark & 56.0 & 67.8 & 53.6 & 46.7 \\
& & \cmark & \textbf{58.6} & \textbf{70.4} & \textbf{55.7} & \textbf{49.9} \\
\bottomrule                       
\end{tabular}
\label{table:benchmark_more_vlm}
\vspace{-10pt}
\end{table}

\noindent \textbf{Comparison with State-of-the-Art Methods on Standard Benchmarks.} Table~\ref{table:benchmark_result} presents the comparison of our approach against state-of-the-art methods. Recently, LLaVA-OneVision~\cite{liLLaVAOneVisionEasyVisual2024a} has achieved strong performance on these datasets, surpassing prior methods such as VideoLLaMA2~\cite{chengVideoLLaMA2Advancing2024} and thus establishing a robust baseline. Our training-free, query-guided inverse transform sampling (ITS) method further enhances the performance of LLaVA-OneVision. Specifically, on Video-MME, accuracy improves from 53.8\% to 56.1\% using 8 frames, and on MLVU, it increases from 58.9\% to 63.4\% with 8 frames. On NextQA, our performance remains comparable to the baseline, likely due to the brevity of the videos. Notably, our method continues to yield improvements even with 32 frames, demonstrating significant gains on long-video benchmarks, such as a 3.2\% increase on LongVideoBench and a 3.6\% increase on MLVU. These results highlight the importance of selecting the right frames at inference time to fully unleash the potential of existing VLMs, and our proposed inverse transform sampling consistently surpasses uniform sampling by a substantial margin.

\vspace{4pt}
\noindent \textbf{Results on Off-the-Shelf VLMs.}
As shown in Table~\ref{table:benchmark_more_vlm}, we extend the proposed approach to additional video VLMs, including Video-LLaVA~\cite{lin2023video}, LongVA~\cite{zhang2024long}, InternVL2~\cite{chen2024internvl}, and LLaVA-Video~\cite{zhang2024video}, and evaluate its effectiveness on the Video-MME benchmark. The results clearly demonstrate that our training-free approach, \methodname{}, consistently boosts the performance of off-the-shelf VLMs, achieving approximately 1\% to 3\% accuracy improvements. For instance, the performance of LLaVA-Video notably increases from 56.0\% to 58.6\% with only 8 frames, verifying the strong generalization capability of inverse transform sampling.

\begin{table}[t!]
\centering
\caption{\textbf{Our BOLT boosts existing VLMs under multi-source retrieval evaluation.} Inverse transform sampling is used. LLaVA-OneVision-7B is utilized and the $K$ is set to 4.}
\vspace{-5pt}
\small
\setlength\tabcolsep{2pt}
\begin{tabular}{lcccc}
\toprule
\multirow{2}{*}{\textbf{Model}}   & \multicolumn{4}{c}{\textbf{Video-MME} \small{(w.o. sub.)}}                         \\\cmidrule(lr){2-5}
                          & Overall     & Short        & Medium      & Long                                   \\ 
\midrule
LLaVA-OneVision$^{\text{64 frame}}$     & 52.1 &  61.8 & 50.8 & 43.8                                        \\
LLaVA-OneVision$^{\text{64 frame}}$ \textbf{+\methodname}  & \textbf{54.9} & \textbf{66.1} & \textbf{54.2} & \textbf{44.4}      \\
\midrule
LLaVA-OneVision$^{\text{16 frame}}$     & 47.8 & 55.8 & 45.6 & 42.0                 \\
LLaVA-OneVision$^{\text{16 frame}}$ \textbf{+\methodname} & \textbf{53.4} & \textbf{65.1} & \textbf{51.1} & \textbf{43.3}         \\
\bottomrule                       
\end{tabular}
\label{table:benchmark_videomme_concact}
\vspace{-12pt}
\end{table}

\vspace{4pt}
\noindent \textbf{Results on Proposed Multi-Source Retrieval Evaluation Setting.} We further evaluate our method under the retrieval-based evaluation setting on Video-MME, as described in Section~\ref{sec:multi_source}. This evaluation assesses the model's ability to retrieve query-relevant information from noisy video content. As shown in Tables~\ref{table:benchmark_result} and~\ref{table:benchmark_videomme_concact}, LLaVA-OneVision's performance significantly decreases from 56.9\% to 47.8\% when using 16 frames, demonstrating that irrelevant background frames substantially hinder comprehension. Even increasing the frame count to 64 frames (equivalent to 16 frames per candidate video) yields only partial recovery (52.1\% vs. 47.8\%), confirming that background noise leads to considerable distraction in pretrained VLMs.

Applying our proposed frame selection strategy consistently improves VLM accuracy. With 16 frames, performance on Video-MME increases from 47.8\% to 53.4\%, representing a substantial 5.6\% gain. Furthermore, results indicate that BOLT effectively enhances accuracy across short, medium, and long videos. These findings highlight both the importance of selecting query-relevant frames and maintaining frame diversity to maximize model performance.

\subsection{Ablations and Analysis}

To further validate the effectiveness of our training-free approach, we conduct ablation studies and analyses from four perspectives: (1) comparisons with alternative frame selection techniques, (2) comparison of visual-text similarity measures, (3) analysis on the hyperparameter $\alpha$, and (4) analysis with respect to video length. Additional experimental results and efficiency analysis are provided in the supplementary material.

\begin{table}[t!]
\centering
\caption{\textbf{Comparison of different training-free frame selection strategies on Video-MME dataset}. LLaVA-OneVision-7B is utilized with 16 frames.}
\vspace{-5pt}
\small
\setlength\tabcolsep{4.2pt}
\begin{tabular}{lccc}
\toprule
\textbf{Frame Selection}    &  \textbf{Query} &  \textbf{Standard} & \textbf{Retrieval-Based}        \\ 
\midrule
Blind Test  & \xmark & \multicolumn{2}{c}{40.85} \\
\midrule
Uniform  & \xmark                             &      56.85     & 47.78       \\
Uniform + Shuffle & \xmark                    &      56.25 (-0.60)    & 46.93 (-0.85)     \\
Random  & \xmark                              &      55.15 (-1.70)    & 46.05 (-1.73)     \\
\midrule
Top-\textit{K}  & \cmark                                & 57.10 (+0.25) & 48.45 (+0.67) \\
Watershed & \cmark                            & 57.23 (+0.38) & 49.15 (+1.37) \\
\textbf{ITS}  & \cmark          & \textbf{57.78} (+0.93) & \textbf{53.37} (+5.59) \\
\bottomrule                       
\end{tabular}
\label{table:ablation_selection_strategy}
\end{table}
\begin{table}[t!]
\centering
\caption{\textbf{Comparison of different visual-text similarity measures on Video-MME dataset}. LLaVA-OneVision-7B is utilized with 16 frames. The baseline is uniform sampling.}
\vspace{-5pt}
\small
\setlength\tabcolsep{6.5pt}
\begin{tabular}{lcccc}
\toprule
\textbf{Baseline}   & \textbf{Caption} & \textbf{SigLIP}    & \textbf{CLIP}              & \textbf{CLIP + SigLIP}                            \\ 
\midrule
56.85 & 55.47 & 57.00 & \textbf{57.78}  & \textbf{57.83} \\
\bottomrule                       
\end{tabular}
\label{table:ablation_vision_encoder}
\vspace{-10pt}
\end{table}

\vspace{4pt} 
\noindent\textbf{Comparison with Alternative Frame Selection Techniques.} Table~\ref{table:ablation_selection_strategy} compares our proposed method with alternative training-free frame selection strategies. As a baseline setup, we exclude visual input entirely, relying solely on the question text (Blind Test). Surprisingly, even without visual data, the model achieves 41\% accuracy on Video-MME, indicating that VLMs heavily utilize general knowledge to answer multiple-choice questions. When visual input is included, uniform and random sampling strategies yield similar performance, around 56\% in standard evaluation and 47\% in multi-source retrieval evaluation, suggesting limited effectiveness of traditional, query-agnostic frame selection approaches on datasets such as Video-MME, which does not require complex temporal reasoning.

However, incorporating query-visual similarity through strategies such as top-\textit{K} sampling, watershed grouping, and inverse transform sampling consistently improves performance over uniform sampling. Among these methods, ITS provides robust gains across both standard and retrieval-based evaluations, achieving the best overall performance. Remarkably, in the retrieval-based evaluation, ITS significantly boosts accuracy from 47.78\% to 53.35\%, highlighting the importance of selecting both query-relevant frames, and underscoring the effectiveness of our proposed method.

\vspace{4pt} 
\noindent\textbf{Comparison of Visual-Text Similarity Measures.} Since our approach leverages similarity between image and query embeddings for frame selection, we compare various image representation methods, including image captions and pretrained vision-language encoders, as shown in Table~\ref{table:ablation_vision_encoder}. We initially caption each frame using LLaVA-OneVision and compute frame selection based on the similarity between queries and generated captions. However, this caption-based approach underperforms due to noisy captions and inaccurate similarity scores. In contrast, directly using image encoders achieves consistent improvements over the uniform sampling baseline, with CLIP outperforming SigLIP. Furthermore, averaging the similarity scores from CLIP and SigLIP yields the best overall performance.

\begin{table}[t!]
\centering
\caption{\textbf{Analysis on the sharpness hyperparameter $\mathbf{\alpha}$ in ITS}. LLaVA-OneVision-7B is utilized with 16 frames.}
\vspace{-5pt}
\small
\setlength\tabcolsep{7pt}
\begin{tabular}{lccc}
\toprule
$\mathbf{\alpha}$   & \textbf{Video-MME} & \textbf{MLVU} & \textbf{LongVideoBench} \\ 
\midrule
Uniform & 56.90 & 61.24 & 55.72\\
\midrule
2   & 58.37 & 65.08 & 58.04 \\
2.5 & 57.51 & 65.24 & 57.00 \\
3   & 57.78 & 65.76 & 57.67 \\
3.5 & 57.74 & 65.11 & 58.41 \\
\bottomrule                       
\end{tabular}
\label{table:ablation_alpha}
\vspace{-4pt}
\end{table}

\vspace{4pt} 
\noindent\textbf{Analysis on the Sharpness Hyperparameter $\alpha$.} As described in Eq.~\ref{eg:alpha}, the hyperparameter $\alpha$ controls the sharpness of the similarity distribution. Table~\ref{table:ablation_alpha} investigates the impact of different $\alpha$ values. When $\alpha$ is set between 2 and 3.5, the VQA performance consistently surpasses uniform sampling across all three datasets by a large margin. Moreover, performance variation within this range is minor, indicating that ITS is less sensitive to the exact value of $\alpha$.

\vspace{4pt} 
\noindent\textbf{Impact of Background Noise Level.} To examine the effect of background noise on model performance, we conduct an analysis using multi-source retrieval evaluation setting, where target videos are concatenated with randomly selected background videos, simulating long and noisy video environments. Table~\ref{table:ablation_video_number} shows that introducing background videos degrades accuracy (frame budget 16), with uniform sampling being particularly susceptible to the increased noise. Although ITS also experiences a performance drop, it consistently outperforms uniform sampling. These results emphasize the effectiveness of query-guided frame selection in mitigating the adverse impacts of noisy contexts.

\begin{table}[t!]
\centering
\caption{\textbf{Impact of video number $K$ in multi-source retrieval evaluation.} If $K\!=\!1$, the evaluation degrades to standard VQA setting. When $K$ is larger, the background noise is also larger.}
\vspace{-5pt}
\small
\begin{tabular}{lcccc}
\toprule
\textbf{Video Number $\bm{K}$} & \textbf{1}   & \textbf{2} & \textbf{4}  & \textbf{8}      \\ 
\midrule
Uniform & 56.85 & 51.81 & 47.78 & 44.82 \\
\midrule
Watershed & 57.23 & 53.70 & 48.67 & 45.89 \\
\textbf{ITS} & \textbf{57.78} & \textbf{54.93} & \textbf{53.37} & \textbf{49.04} \\
\bottomrule                       
\end{tabular}
\label{table:ablation_video_number}
\vspace{-8pt}
\end{table}

\subsection{Visualization}
\begin{figure}[t]
\centering
\includegraphics[width=0.95\linewidth]{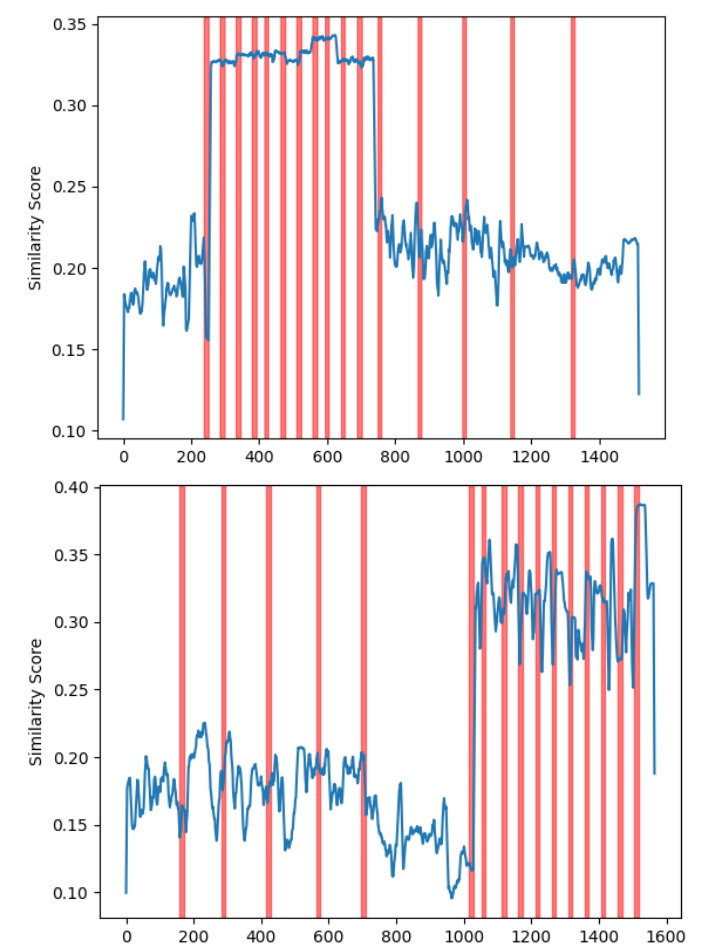}
\vspace{-6pt}
\caption{\textbf{Visualization of frame selection results using inverse transform sampling.} The blue curve represents the similarity scores across the entire video sequence, and the red lines indicate the selected frames.}
\label{fig:visual}
\vspace{-10pt}
\end{figure}

To further illustrate the frame selection results of inverse transform sampling, we provide two examples from the Video-MME in Figure~\ref{fig:visual}. As shown in the figure, ITS can effectively sample query-relevant frames with high similarity scores. At the same time, ITS also includes certain local peaks as contextual information, ensuring frame diversity and providing the VLMs with crucial context for reasoning.

\section{Conclusion}

In this paper, we introduced BOLT, a training-free method designed to enhance large VLMs for long-form video understanding. We explored various temporal frame selection techniques, focusing particularly on inverse transform sampling, to efficiently prioritize query-relevant frames, significantly improving visual question-answering performance. Our evaluations demonstrate that BOLT effectively addresses the limitations of conventional uniform sampling, providing a scalable solution to extend the capabilities of existing VLMs without requiring additional fine-tuning.

\noindent\textbf{Limitations.} Although BOLT significantly improves long-form video understanding, its effectiveness relies on accurate query-frame similarity measures, which may not always reflect true relevance on tasks that require reasoning. Future work could explore more robust similarity metrics that better adapt to varied video content and query contexts, including under multi-round VQA scenarios.

\noindent\textbf{Acknowledgments.} This work is supported by the KAUST Center of Excellence for Generative AI under award number 5940. The computational resources are provided by IBEX, which is managed by the Supercomputing Core Laboratory at KAUST.

{
    \small
    \bibliographystyle{ieeenat_fullname}
    \bibliography{main}
}

\renewcommand\thesection{\Alph{section}}
\renewcommand\thesubsection{\thesection.\arabic{subsection}}
\setcounter{section}{0}
\newpage
\maketitlesupplementary

In this appendix, we provide additional experimental results and further analyses. Specifically, we present results on Video-MME with subtitles in Section~\ref{sec:videomme_subtitle}. Then, we discuss inference costs in Section~\ref{sec:supp_cost}. In the end, we provide additional visualization examples in Section~\ref{sec:supp_visual}.

\section{Additional Results on Video-MME}
\label{sec:videomme_subtitle}

In the main paper, we presented benchmark results of various VLMs on the Video-MME dataset without subtitles. To further validate the effectiveness of the proposed method, we incorporate subtitles into the VLMs’ text input. As shown in Table~\ref{table:videomme_subtitle}, our method consistently improves overall performance across different frame budgets. 
Particularly, with only 8 input frames, BOLT increases the accuracy from 58.7\% to 61.0\%. The performance across short, medium, and long videos also improves, demonstrating the effectiveness and robustness of our approach in leveraging both visual and textual information.

\begin{table}[h!]
\centering
\caption{\textbf{Benchmark results on Video-MME dataset with subtitles.} Our proposed inverse transform sampling can consistently enhance the overall performance under different frame budgets.}
\small
\setlength\tabcolsep{1.8pt}
\begin{tabular}{lccccc}
\toprule
\textbf{Model} & \textbf{BOLT}  & \textbf{Overall}     & \textbf{Short}        & \textbf{Medium}      & \textbf{Long}                            \\ 
\midrule
\multirow{2}{*}{LLaVA-OneVision$^{\text{8 frame}}$} &\xmark & 58.7 & 70.8 & 56.1 & 49.4  \\
& \cmark & \textbf{61.0} & \textbf{71.7} & \textbf{58.4} & \textbf{52.9}     \\
\midrule
\multirow{2}{*}{LLaVA-OneVision$^{\text{16 frame}}$} &\xmark & 60.3 & 72.7 & 57.1 & 51.1  \\
& \cmark  & \textbf{61.7} & \textbf{74.0} & \textbf{59.3} & \textbf{51.8}     \\
\midrule
\multirow{2}{*}{LLaVA-OneVision$^{\text{32 frame}}$} &\xmark & 61.9 & \textbf{75.7} & 58.4 & \textbf{51.6}  \\
& \cmark  & \textbf{62.7} & 75.1 & \textbf{61.4} & 51.4     \\
\bottomrule                     
\end{tabular}
\label{table:videomme_subtitle}
\vspace{-4pt}
\end{table}

\section{Analysis of Inference Cost}
\label{sec:supp_cost}

We also evaluate the inference cost of our training-free approach. Our method consists of three main steps: CLIP-based frame feature encoding, inverse transform sampling, and VLM inference. In terms of memory usage, our approach requires nearly the same GPU memory as the baseline method that uses uniform sampling. In terms of inference time, as shown in Table~\ref{table:time}, the total inference time per sample increases by approximately 90\%, primarily due to the CLIP feature encoding step. In contrast, the inverse transform sampling itself is highly efficient. Although our method introduces some additional inference time, it remains acceptable considering that no training or fine-tuning is required.

\begin{table}[h!]
\centering
\caption{\textbf{Inference time analysis.} The inference time is evaluated by one A100 GPU. We utilize the LLaVA-OneVision-7B with an input of 16 frames. CLIP-L/14 is used to extract visual features.}
\small
\begin{tabular}{lrr}
\toprule
\textbf{Step} & \textbf{Time} & \textbf{Increase} \\
\midrule
VLM inference & 1.32 s \\             
\midrule
CLIP visual feature & 1.21 s \\
Inverse Transform Sampling & 0.003 s \\
\midrule
Total &  2.53 s & +90.9\% \\
\bottomrule                       
\end{tabular}
\label{table:time}
\end{table}

Additionally, our inference pipeline relies solely on basic CLIP for visual-text matching. While incorporating auxiliary alignment models or external tools, such as object detectors or OCR models, could further improve VQA performance, it would inevitably increase computational overhead. Our frame selection method is orthogonal to such approaches, as it focuses on selecting query-relevant frames to improve the effectiveness of downstream VLM inference.

\begin{figure}[b]
\centering
\includegraphics[width=0.94\linewidth]{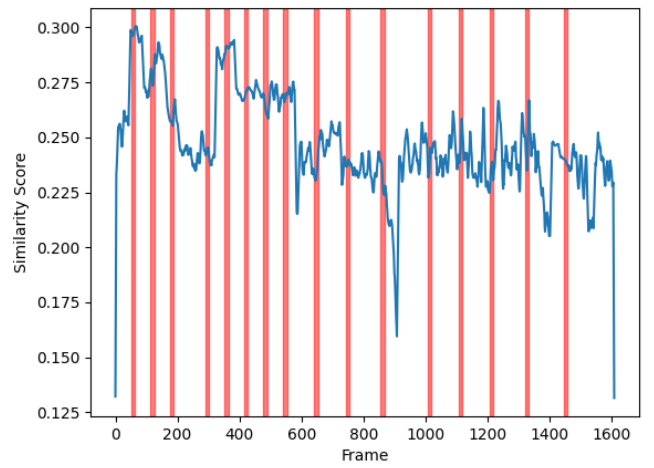}
\vspace{-5pt}
\caption{\textbf{When visual-query similarity scores remain similar across the video, inverse transform sampling simplifies to uniform sampling.}}
\label{fig:supp_visual_f}
\end{figure}

\section{Additional Qualitative Results}
\label{sec:supp_visual}

\begin{figure}[t]
\centering
\includegraphics[width=0.94\linewidth]{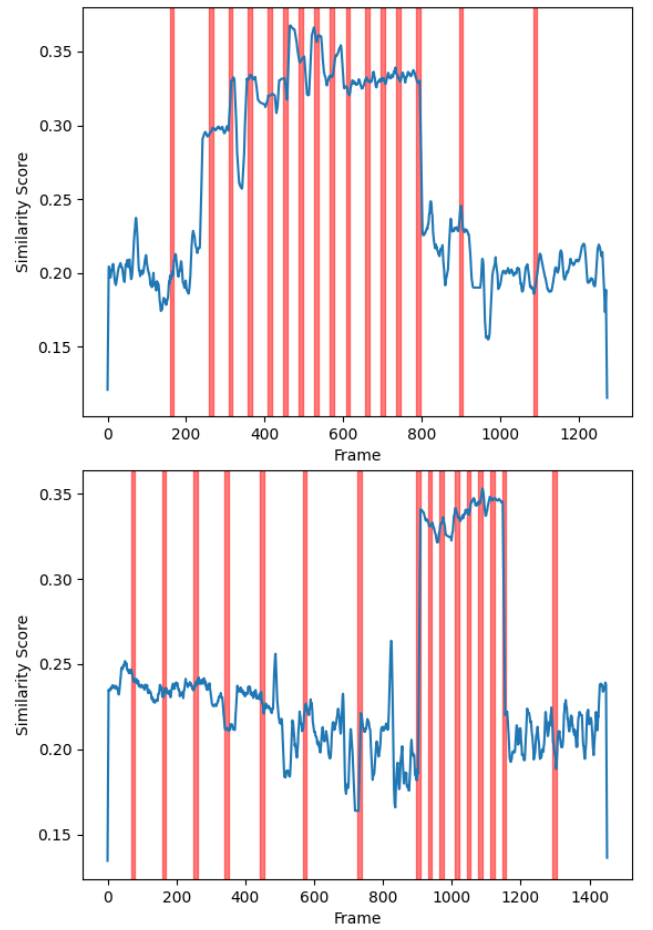}
\vspace{-5pt}
\caption{\textbf{Additional visualization results.}}
\label{fig:supp_visual}
\vspace{-5pt}
\end{figure}

We provide additional visualizations of inverse transform sampling in Figure~\ref{fig:supp_visual}. The blue curve represents the similarity scores across the entire video sequence, while the red lines indicate the selected frames.

As shown in the figure, the proposed method effectively selects frames with high visual-query similarity. In addition, it preserves certain background elements, helping to maintain important contextual information required for accurate video understanding.

In Figure~\ref{fig:supp_visual_f}, we present an example where the visual-query similarity scores remain relatively similar throughout the video. In such cases, the cumulative distribution function becomes approximately linear, causing the final selected frames to approximate uniform sampling. This indicates that most clips in the video may contribute equally to answering the question.

\end{document}